\documentclass[letterpaper, 10 pt, conference]{template/ieeeconf}  

\IEEEoverridecommandlockouts                              

\usepackage{amsmath} 
\usepackage{amssymb}  
\usepackage{graphicx}
\usepackage[caption=false]{subfig}
\usepackage{amsmath} 
\usepackage{amssymb}  
\usepackage{hyphenat}
\usepackage[table]{xcolor}
\usepackage[font=small,belowskip=0pt, aboveskip=6pt]{caption}
\usepackage{multirow}
\usepackage{url}
\usepackage{booktabs,amsfonts,dcolumn}
\usepackage{tabularx}
\usepackage{tabu}
\usepackage{gensymb}
\usepackage{makecell}
\usepackage{hyperref}
\usepackage{comment}
\usepackage[style=ieee,mincitenames=1,maxcitenames=2,maxbibnames=10,doi=false,isbn=false,url=false,eprint=false]{biblatex}
\DeclareSourcemap{
  \maps[datatype=bibtex, overwrite]{
    \map{
      \step[fieldset=address, null]
      \step[fieldset=location, null]
    }
  }
}

\usepackage{xspace}

\long\def\invis#1{}

\newcommand\eq[1]{Eq.~\eqref{#1}}
\newcommand\fig[1]{Fig.~\ref{#1}}

\makeatletter
\DeclareRobustCommand\onedot{\futurelet\@let@token\@onedot}
\def\@onedot{\ifx\@let@token.\else.\null\fi\xspace}

\makeatother

\usepackage[group-separator={,}]{siunitx}

\title{\LARGE \bf
EDI: ESKF-based Disjoint Initialization for Visual-Inertial SLAM Systems 
 }

\author{Weihan Wang$^{a}$, Jiani Li$^b$, Yuhang Ming$^c$, Philippos Mordohai$^a$
\thanks{$^a$Stevens Institute of Technology, Hoboken, NJ, USA, 07030, {\tt\small \{wwang103,pmordoha\}@stevens.edu}}
\thanks{$^b$Meta, {\tt\small jiani.li@meta.com}}
\thanks{$^c$School of Computer Science, Hangzhou Dianzi University, Hangzhou 310018, China, 
{\tt\small yuhang.ming@ieee.org.}}%
\thanks{
This research has been supported in part by the National Science Foundation under award 2024653.} %
}

\begin{document}

\begin{minipage}{0.90\textwidth}\ \\[12pt]
\vspace{3in}
\begin{center}
     This paper has been accepted for publication in \textit{IEEE/RSJ International Conference on Intelligent Robots 2023}.  
\end{center}
  \vspace{1in}
  ©2023 IEEE. Personal use of this material is permitted. Permission from IEEE must be obtained for all other uses, in any current or future media, including reprinting/republishing this material for advertising or promotional purposes, creating new collective works, for resale or redistribution to servers or lists, or reuse of any copyrighted component of this work in other works.
\end{minipage}

\newpage

\maketitle
\thispagestyle{empty}
\pagestyle{empty}

\begin{abstract}
Visual-inertial initialization can be classified into joint and disjoint approaches. Joint approaches tackle both the visual and the inertial parameters together by aligning observations from feature-bearing points based on IMU integration then use a closed-form solution with visual and acceleration observations to find initial velocity and gravity. In contrast, disjoint approaches independently solve the Structure from Motion (SFM) problem and determine inertial parameters from up-to-scale camera poses obtained from pure monocular SLAM. However, previous disjoint methods have limitations, like assuming negligible acceleration bias impact or accurate rotation estimation by pure monocular SLAM. To address these issues, we propose EDI, a novel approach for fast, accurate, and robust visual-inertial initialization. Our method incorporates an Error-state Kalman Filter (ESKF) to estimate gyroscope bias and correct rotation estimates from monocular SLAM, overcoming dependence on pure monocular SLAM for rotation estimation. To estimate the scale factor without prior information, we offer a closed-form solution for initial velocity, scale, gravity, and acceleration bias estimation. To address gravity and acceleration bias coupling, we introduce weights in the linear least-squares equations, ensuring acceleration bias observability and handling outliers. Extensive evaluation on the EuRoC dataset shows that our method achieves an average scale error of 5.8\% in less than 3 seconds, outperforming other state-of-the-art disjoint visual-inertial initialization approaches, even in challenging environments and with artificial noise corruption.
\end{abstract}

\section{INTRODUCTION}
The combination of a single camera and an Inertial Measurement Unit (IMU) in Visual-Inertial Navigation Systems (VINS) is a cost-efficient and low-power solution for robot perception and AR/VR applications. The camera provides a rich representation of the environment while the IMU measures acceleration and angular velocity, making it robust to fast-motion and texture-less images. This combination makes them ideal for complementing each other. 
\begin{figure}[!ht]
    \vspace{0.05in}
    \begin{center}
\includegraphics[width=\columnwidth]{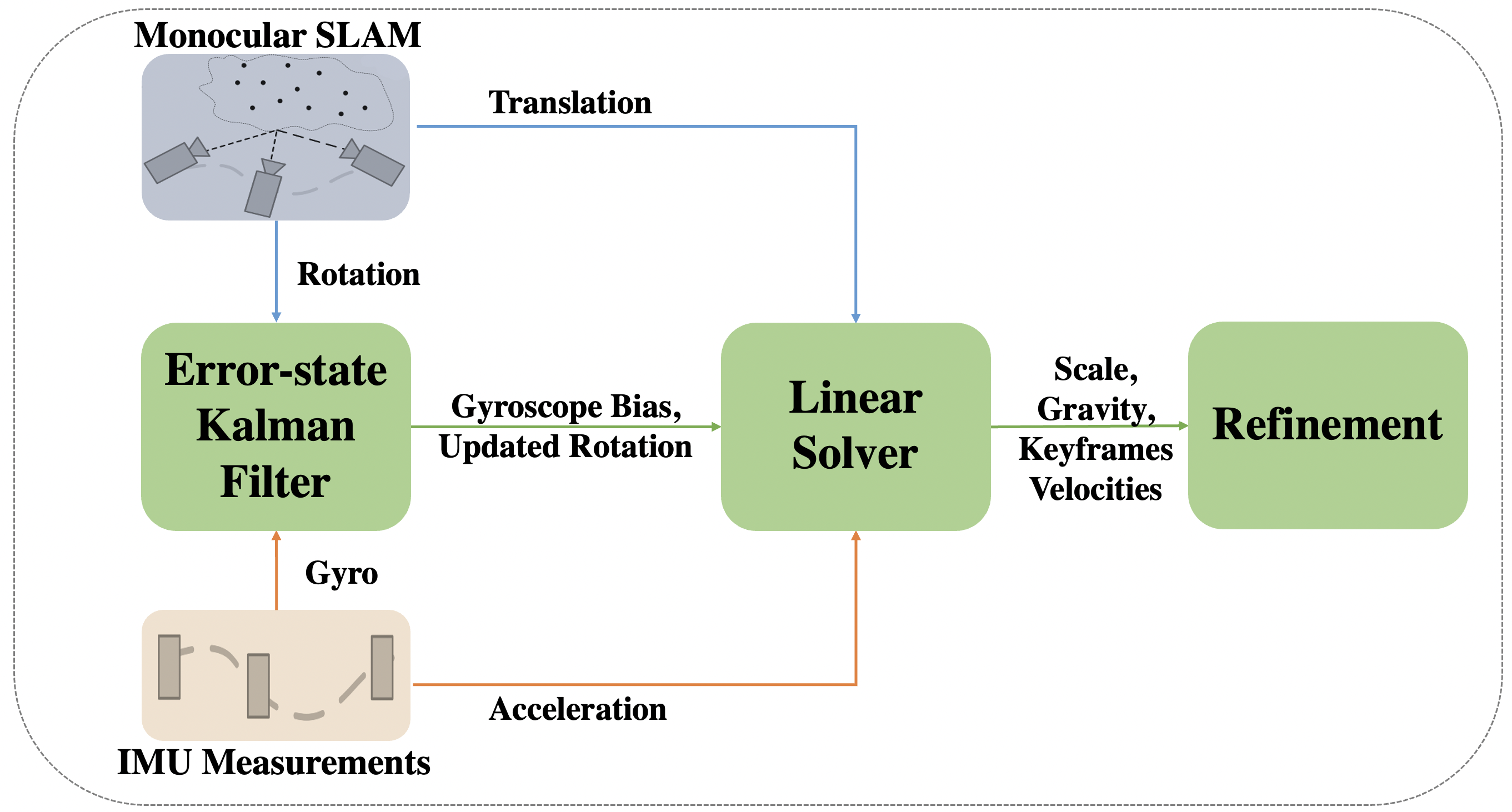}
    \end{center}
    \caption{A diagram of the components of EDI. The green blocks represent steps 1 to 3 of EDI, and the different colored arrows represent different output flows from their corresponding steps.}
     
    \label{fig:diagram}
    \vspace{-0.07in}
\end{figure}
The initialization process is crucial for VINS, as it requires a good initial estimation for the scale, gravity, initial velocity, acceleration, and gyroscope biases, but remains a challenge as it requires fast and accurate recovery of observable parameters  from visual and inertial measurements without prior knowledge. Poor initialization can lead to trajectory drift and hinder the convergence of subsequent optimization. A prolonged initialization process is also impractical for both robotics and AR/VR applications. 

\begin{table*}[!ht]
\vspace{0.20in}
\centering
\caption{Assumptions underlying joint and disjoint initialization methods. The fewer assumptions the methods make, the more versatile and practical they become.}
\scalebox{0.85}{
\begin{tabular}{@{}clcccccc}
\toprule
 &Methods& \makecell[c]{Noiseless \\ IMU
and camera \\ measurement}& \makecell[c]{All features\\ tracked in\\ all frames} &\makecell[c]{Acceleration
bias \\is negligible}& \makecell[c]{Gyroscope
bias \\is negligible}& \makecell[c]{Known Camera-IMU\\ extrinsic calibration}&\makecell[c]{Monocular SLAM \\accurately estimates\\ camera pose}\\
\midrule
\parbox[t]{2mm}{\multirow{5}{*}{\rotatebox[origin=c]{90}{Joint}}} & Kneip et al~\cite{Kneip}& $\surd$&$\surd$&$\surd$&$\surd$&$\surd$&$\times$\\
&Martinelli~\cite{Martinelli}& $\surd$&$\surd$&$\times$&$\surd$&$\surd$&$\times$\\
&Kaiser et al~\cite{Kaiser}& $\surd$&$\surd$&$\surd$&$\times$&$\surd$&$\times$\\
&Dong-Si and Mourikis~\cite{DongSi},~\cite{DongSi2}& $\surd$&$\surd$&$\times$&$\times$&$\times$&$\times$\\
&Campos et al.~\cite{jointviinit}& 
$\surd$&$\times$&$\times$&$\times$&$\surd$&$\times$\\
 \midrule
\parbox[t]{2mm}{\multirow{5}{*}{\rotatebox[origin=c]{90}{Disjoint}}}&ORB-SLAM-VI~\cite{viorb}& $\times$&$\times$&$\times$&$\times$&$\surd$&$\surd$\\
&Huang and Liu~\cite{weibovio} & $\times$&$\times$&$\times$&$\times$&$\times$&$\surd$\\
&VINS-Mono~\cite{VINS}& $\times$&$\times$&$\surd$&$\times$&$\surd$&$\surd$\\
&ORB-SLAM3~\cite{ORBSLAM3TRO}& $\times$&$\times$&$\times$&$\times$&$\surd$&$\surd$\\
&EDI (Ours)& $\times$&$\times$&$\times$&$\times$&$\surd$&$\times$\\
\bottomrule
\end{tabular}
}
\label{tab:assumption_summary}
\end{table*}

Visual-inertial initialization is classified into two categories~\cite{David},~\cite{jun2021improved}: joint~\cite{Kneip}-\cite{jointviinit} and disjoint approaches~\cite{viorb},~\cite{weibovio},~\cite{VINS},~\cite{inertialonlyinit}.
Joint approaches tackle both the visual and the inertial parameters together by aligning observations from feature-bearing points based on IMU integration.  
Joint visual-inertial initialization methods begin by finding a closed-form solution to the visual-inertial problem.   
A closed-form solution to calculate the initial velocity and position of feature points using only the visual information from three consecutive image frames and a single feature point was first introduced by Kneip et al.~\cite{Kneip}. Martinelli~\cite{Martinelli} later proposed a closed-form solution that also takes into account the scale, gravity, acceleration bias and feature points' depth, and analyzes the necessary conditions for the solution to be attainable. Kaiser et al.~\cite{Kaiser} apply Martinelli's solution and utilize non-linear optimization to account for the impact of gyroscope bias on the system.
These methods track multiple points in all images and use a system of equations to minimize the 3D error of feature points in space. 
Dong-Si and Mourikis~\cite{DongSi},~\cite{DongSi2} proposed a closed-form solution for estimating the attitude, velocity, feature positions, and camera-IMU extrinsic calibration. In their work, they also specifically discuss two methods for recovering the relative rotation between the camera and IMU under different scenarios with varying numbers of tracked features.
Campos et al.~\cite{jointviinit} further improve joint methods by leveraging preintegration to reduce the computational cost of the closed-form initialization and conducting two rounds of visual-inertial Bundle Adjustment (VI-BA) to increase the precision of depth feature estimates, gyroscope bias, gravity, and initial velocity.
However, these joint methods are based on the assumption that there is no noise in the IMU and camera measurements, and that all feature points are tracked correctly in all frames. Even though the work by Campos et al.~\cite{jointviinit} relaxes the requirement for tracking feature points in all frames, it still faces challenges with low recall rate, leading to extended initialization times.

In contrast, disjoint approaches aim to solve the SfM problem independently first, and then determine the inertial parameters based on up-to-scale camera poses obtained from a pure monocular simultaneous localization and mapping (SLAM) system. This approach is made possible by the use of monocular SLAM, which performs local bundle adjustment and takes into consideration both photometric and geometric consistency in feature point tracking, providing a more precise state estimation.
One method in this category, used in ORB-SLAM-VI~\cite{viorb}, is proposed by Mur-Artal and Tardós. It runs monocular SLAM for a few seconds, assuming the sensor undergoes a motion that makes all variables observable and divides the initialization process into four sub-problems. 

Later, Huang and Liu~\cite{weibovio} adopted this concept and expanded on it by incorporating an estimate of the camera-IMU extrinsic parameters.
Qin et al. in VINS-Mono~\cite{VINS} align the camera's trajectory and orientation with the IMU preintegration measurements by solving linear least-squares equations to obtain values for the scaled velocity, scale, gravity, and gyroscope biases, resulting in  high-performance output compared to other state-of-the-art visual-inertial odometry systems such as OKVIS\cite{okvis}, SVO~\cite{svo}, and ROVIO~\cite{ROVIO}, as evaluated on various datasets~\cite{ExperimentalComparisonvio}.
Both methods determine inertial parameters by solving a set of linear equations using least-squares, but they differ in the steps involved. ORB-SLAM-VI does not take velocity into account, while VINS-Mono does not estimate the acceleration bias when solving a set of linear equations.
However, both methods have limitations. For example, ORB-SLAM-VI requires 15 seconds of initialization to make the acceleration bias observable, and both methods assign equal weights to the residues without considering IMU measurement uncertainty when solving linear least-squares. To address these limitations in the above disjoint methods, Campos et al.~\cite{inertialonlyinit} proposed a new disjoint initialization method in ORB-SLAM3~\cite{ORBSLAM3TRO} by formulating the visual-inertial initialization as a maximum a-posteriori (MAP) problem and demonstrate that their method outperforms the best joint and previous disjoint initialization methods.
However, this method is sensitive to the scale factor. The second step in this approach, which involves the iterative inertial-only optimization to estimate the scale factor, requires a reliable initial estimate of scale.
Furthermore, the method requires an empirical prior residual for the biases. Although the third step of the ORB-SLAM3 initialization phase, which involves joint visual-inertial bundle adjustment (VI-BA), aims to improve the previous estimate from the second step, the quality of the second step can greatly impact the convergence time of the third step and the ability of the VI-BA to produce an optimal solution. Additionally, the success of the disjoint method is highly dependent on the performance of the pure monocular SLAM system.

To overcome the limitations of the previous disjoint methods, we propose EDI, an innovative disjoint approach for initializing a visual-inertial system.
The main contributions of the  proposed initialization method are:
\begin{itemize}
    \item Eliminating the need for computationally intensive visual-inertial bundle adjustment (VI-BA) while ensuring accuracy, improving the efficiency of the method, and increasing robustness in challenging conditions.
    \item Proposing a new method that incorporates an Error-state Kalman Filter (ESKF)~\cite{Joan} to estimate gyroscope bias and correct rotation estimates with monocular SLAM, considering the probabilistic model of IMU noise. 
    \item  Providing a closed-form solution for estimating initial velocity, scale, gravity, and acceleration bias, along with weights to handle outliers.
\end{itemize}

 The summary of assumptions underlying joint and disjoint initialization methods, including ours, are shown in Table~\ref{tab:assumption_summary}.

\section{PRELIMINARIES}
\subsection{Notation}\label{sec:notation}
In this paper,  the world frame is represented by $(\cdot)^\text{w}$, the body frame is represented by $(\cdot)^\text{b}$, and the camera frame is represented by $(\cdot)^\text{c}$.  $(\hat{\cdot})$ represents a posterior estimate after being corrected by ESKF, while an up-to-scale estimate is denoted as $(\bar{\cdot})$ . We use $\textbf{R}$ for rotation matrices and $\textbf{g}$ for the gravity vector. The rotation, translation and velocity from the body frame to the world frame are represented by $\textbf{R}^{\text{w}}_{\text{b}}$, $\textbf{p}^{\text{w}}_{\text{b}}$ and $\textbf{v}^{\text{w}}_{\text{b}}$ respectively. $\text{b}_k$ is the body frame while taking the $k$-th image, and $\text{c}_k$ is the camera frame while taking the $k$-th image. Acceleration bias and gyroscope bias in the local body frame are represented by $\textbf{b}_\text{a}$ and $\textbf{b}_\text{g}$ respectively. The nominal state vectors are represented by $(\cdot)$, true state vectors are represented by $(\cdot)_t$, the error-state is represented by $\delta(\cdot)$ and $\mathcal{X}$ is a state variable.


\subsection{The Error-State Kalman Filter}
The Error-state Kalman filter, or ESKF~\cite{Joan}, is a type of filter used to estimate the true state of a system while taking into account measurement noise and model uncertainty. It is widely used in control systems, navigation and signal processing. It offers advantages such as minimal representation of state variables in rotation processing, and operating near the origin to avoid linearization approximation issues and gimbal lock problems. The state variables in ESKF are minimal, allowing for the omission of second-order variables, and the Jacobian matrices 
are also straightforward and can even be substituted with identity matrices.

In ESKF, there are three state variables: true, nominal, and error state. 
The nominal state 
integrates with noise and other potential model flaws, leading to the accumulation of errors and drift. The error state accounts for various noise sources and biases. The relationship among the true state ($\mathcal{X}_t$), the nominal state ($\mathcal{X}$) and the error state ($\delta\mathcal{X}$) is defined as
\begin{equation*}\small \label{generic_compostion}
\mathcal{X}_t  = \mathcal{X}  \oplus \delta\mathcal{X}
\end{equation*}
where $\oplus$ indicates a generic composition.



The procedure of the ESKF outlined in this paper is as follows: upon receipt of an IMU measurement, it is integrated and incorporated into the nominal state variables. The error state takes into account the noise term and biases in the ESKF, providing a Gaussian distribution for the error state. The ESKF also incorporates a prediction and correction process, utilizing observations from sensors other than the IMU. Following correction, the ESKF yields a posterior error Gaussian distribution, and the error is incorporated into the nominal state variables, resetting the ESKF. This process is repeated iteratively.

\section{Proposed Approach}\label{sec:method}
This section describes the proposed online initialization method, EDI, aiming to estimate precise initial values for  the body velocities, gravity direction, scale factor, gyroscope and acceleration bias. 


We begin by discussing the techniques used in two state-of-the-art disjoint initialization methods and comparing them to our own. ORB-SLAM3~\cite{ORBSLAM3TRO} uses a three-step process: vision-only MAP estimation, inertial-only MAP estimation, and  joint visual-inertial optimization for further refining the solution. 
While this method can provide accurate results, it is computationally demanding due to the extensive non-linear optimization procedure. On the other hand, VINS-Mono~\cite{VINS}, a linear loosely-coupled initialization method, is quick to compute but relies on the assumption of zero acceleration bias.

Our method offers the best of both worlds - it is as efficient as VINS-Mono and produces an estimation comparable to ORB-SLAM3 without the need for multiple non-linear optimization steps or strong assumptions. Furthermore, it is more robust in challenging conditions.
Our method, as shown in \fig{fig:diagram}, is composed of four steps: 
\begin{itemize}
    \item \textbf{Step 0. Pure Monocular SLAM}: Obtain the initial keyframe poses with an unknown scale.
    \item \textbf{Step 1. ESKF-based Gyroscope Bias Estimation}: Utilize the ESKF to estimate the gyroscope bias and combine rotation estimates from IMU prediction and pure monocular SLAM.
    \item \textbf{Step 2. Linear Solver}: Align the IMU trajectory and pure monocular SLAM trajectory by determining the scale factor, keyframes' velocities, and gravity.
    \item \textbf{Step 3. Refinement}: 
Use the solution from the previous step as the initial estimate to obtain 
acceleration bias, refined scale, keyframes' velocities and gravity estimate.
\end{itemize}
As a disjoint method, EDI uses a pure monocular SLAM \cite{mur2015orb} with an increased keyframe insertion rate for a short period of time (1 or 2 seconds) to obtain the initial keyframe poses with an unknown scale, in order to ensure observability of all inertial variables.

\subsection{ESKF-based Gyroscope Bias Estimation}\label{sec:eskf}

To estimate the gyroscope bias and combine rotation estimates from the IMU prediction and monocular SLAM during the initialization stage with a window of $N$ keyframes, we only consider the rotation and gyroscope bias in the nominal state and error state in this step.

Consider two consecutive keyframes $\text{b}_{k}$ and $\text{b}_{k+1}$ with interval $\Delta t$ and denote the nominal state at keyframe $\text{b}_{k+1}$ as $\mathcal{X}_{\text{b}_{k+1}} = \left [ \textbf{R}^{\text{w}}_{\text{b}_{k+1}}, \textbf{b}_{\text{g}, \text{b}_{k+1}} \right]$ and the error state as $\delta\mathcal{X}_{\text{b}_{k+1}} = \left [ \delta\boldsymbol\theta_{\text{b}_{k+1}}, \delta\textbf{b}_{\text{g}, \text{b}_{k+1}} \right]$. The error state of rotation and gyroscope bias is integrated, allowing the IMU measurements to make predictions for the ESKF as follows:
\begin{equation}\small\label{eq:theta_equation}
\delta \boldsymbol\theta_{\text{b}_{k+1}}= \text{Exp}(-(\boldsymbol{\omega}_m- \textbf{b}_g)\Delta t)\delta {\boldsymbol{{\theta}}}_{\text{b}_{k}} - \delta {\textbf{b}}_{\text{g}, \text{b}_{k}}\Delta t + \boldsymbol\eta_{\theta},
\end{equation}
\begin{equation}\small\label{eq:e_gyroscope_equation}
\delta \textbf{b}_{\text{g}, \text{b}_{k+1}} = \delta {\textbf{b}}_{\text{g}, \text{b}_{k}} +  \boldsymbol\eta_{g},
\end{equation}
where $\boldsymbol{\omega}_m$ is the raw gyroscope measurement,
$\delta{\boldsymbol{{\theta}}}_{\text{b}_{k}}$ and $\delta {\textbf{b}}_{\text{g}, \text{b}_{k}}$ are  error state of keyframe $\text{b}_{k}$,
$\boldsymbol\eta_{\theta}$ and $\boldsymbol\eta_{g}$ are white Gaussian noise applied to rotation and gyroscope bias estimation respectively, $\boldsymbol\eta_{\theta} \sim \mathcal{N}(0, \sigma^2_{w_n}\Delta t^2\textbf{I})$, $\boldsymbol\eta_{g} \sim \mathcal{N}(0, \sigma^2_{w_w}\Delta t\textbf{I})$.
Meanwhile, the covariance matrix of the prediction at $\text{b}_{k+1}$ is updated as follows:
\begin{equation*}\small
\boldsymbol{P}_ {{\text{pred}}} =   \boldsymbol{F}\hat{\boldsymbol{P}}_{\text{b}_{k}}\boldsymbol{F}^\top + \boldsymbol{Q},
\end{equation*}
where $\boldsymbol{P}_ {{\text{pred}}}$ is the predicted covariance matrix of keyframe $\text{b}_{k+1}$, $\hat{\boldsymbol{P}}_{\text{b}_{k}}$ is the corrected covariance matrix of keyframe $\text{b}_{k}$, 
$\boldsymbol{Q}$ is the covariance matrix of the perturbation impulses~($\boldsymbol{Q} = \text{diag}(\text{Cov}(\boldsymbol\eta_{\theta}), \text{Cov}(\boldsymbol\eta_{g}))$), and $\boldsymbol{F}$ is the Jacobian matrix with respect to the error state of keyframe $\text{b}_{k}$ based on \eq{eq:theta_equation} and \eq{eq:e_gyroscope_equation}:
\begin{equation*}\small
\begin{split}
\boldsymbol{F} = \left [ 
\begin{matrix}
\text{Exp}(-(\boldsymbol{\omega}_m- \textbf{b}_g)\Delta t) & -I\Delta t \\
0 & I
\end{matrix}
 \right].
\end{split}
\end{equation*}

In order to avoid drift in the IMU prediction from integrating the IMU measurement directly, we need to correct  the IMU prediction with other complementary sensors. In this paper,  we treat monocular SLAM as a sensor and use its rotation estimates as observations to fuse with  our IMU predictions. We imagine this abstract sensor as a typical sensor that gives information based on the current state:
\begin{equation*}\small
\boldsymbol{r}^\text{w}_{\text{b}_{k+1}} = \boldsymbol{h}(\mathcal{X}_t) + \boldsymbol{v},
\end{equation*}
where $\boldsymbol{r}^\text{w}_{\text{b}_{k+1}}$ is an orientation observation at keyframe $\text{b}_{k+1}$ from pure monocular SLAM, $\boldsymbol{h}()$ is the observation function of the system, $\boldsymbol{v}$ is white Gaussian noise with covariance $\boldsymbol{V}$, $\boldsymbol{v} \sim \mathcal{N}(0, \boldsymbol{V})$ and the orientation difference between prediction and observation is denoted as $\textbf{e}_r 
~(\textbf{e}_r = \text{Log}(\boldsymbol{h}({\mathcal{X}_t})^{\top}\boldsymbol{r}^{\text{w}}_{\text{b}_{k+1}}))$.
In ESKF, our goal is to update the error state, so we need to calculate the Jacobian matrix $\boldsymbol{H}$, which is the matrix of partial derivatives of the observation with respect to the error state $\delta \mathcal{X}$:
\begin{equation*}\small
\boldsymbol{H} = \frac{\partial{\boldsymbol{h}}}{\partial{\delta \mathcal{X}_{\text{b}_{k+1}}}} =  \frac{\partial{\boldsymbol{h}}}{\partial{\mathcal{X}_t}}\frac{\partial{\mathcal{X}_t}}{\partial{\delta \mathcal{X}_{\text{b}_{k+1}}}} = \boldsymbol{J}_r^{-1}(\textbf{e}_r ), \\
\end{equation*}
where $\boldsymbol{J}_r^{-1}$ is the inverse of the right Jacobian.

Then, we calculate the Kalman gain and the update 
of the error state as follows:
\begin{equation*}\small
\boldsymbol{K} = \boldsymbol{P}_{\text{pred}}\boldsymbol{H}^{\top}(\boldsymbol{H}\boldsymbol{P}_{{\text{pred}}}\boldsymbol{H}^{\top} + \boldsymbol{V})^{-1},
\end{equation*}
\begin{equation*}\small
\delta \hat{\mathcal{X}}_{\text{b}_{k+1}} =  \boldsymbol{K}(\text{Log}(\textbf{e}_r)),
\end{equation*}
\begin{equation*}\small
\hat{\boldsymbol{P}}_{\text{b}_{k+1}} = (\boldsymbol{I} - \boldsymbol{K}\boldsymbol{H})\boldsymbol{P}_{{\text{pred}}}.
\end{equation*}
where $\boldsymbol{K}$ is the Kalman gain.

After the ESKF update, the posterior error state $\delta \hat{\mathcal{X}}_{\text{b}_{k+1}}$, is incorporated into the nominal state. Afterwards, $\delta \hat{\mathcal{X}}_{\text{b}_{k+1}}$ is reset to zero, and its corresponding covariance matrix is updated to reflect this reset accordingly. The best true-state estimate at keyframe $\text{b}_{k+1}$ is obtained using the appropriate compositions as follows: 
\begin{equation*}\small
\begin{split}
{\hat{\textbf{R}}^{\text{w}}_{\text{b}_{k+1}}} = {\textbf{R}^{\text{w}}_{\text{b}_{k+1}}}\text{Exp}(\delta \hat{\boldsymbol\theta}_{\text{b}_{k+1}}), \\
\hat{\textbf{b}}_{\text{g}, \text{b}_{k+1}}= \textbf{b}_{\text{g}, \text{b}_{k+1}} + \delta \hat{\textbf{b}}_{\text{g}, \text{b}_{k+1}}.
\end{split}
\end{equation*}

Initially, the gyroscope bias is assumed to be zero. By using the ESKF with $N$ keyframes, the estimated gyroscope bias
in the last keyframe in the window is considered the most accurate true-state estimate. We use this as the final estimated gyroscope bias. Additionally, our method is different from other initialization methods in that it fuses the rotation estimate from the IMU prediction with the rotation estimate from monocular SLAM for each keyframe within the window. This allows for a more accurate estimation of the rotation when monocular SLAM is not accurate and robust enough.

\subsection{Linear Solver}\label{sec:linear}
This step aims to obtain an optimal estimate of the keyframes' velocities, gravity and scale factor of the pure monocular SLAM.  As the pure monocular SLAM system \cite{mur2015orb} only estimates keyframe poses without recovering the scale, the estimation of acceleration bias, correction of keyframe translations, velocities and gravity using ESKF is not possible. To overcome this limitation, we solve a set of linear equations to obtain the following estimates:
\begin{equation*}\small
\mathcal{X}_1 = \left [\textbf{v}^{\text{w}}_{\text{b}_{0}:\text{b}_{N-1}},\textbf{g}^{\text{w}}, \text{s}\right]^\top.
\end{equation*}

Considering two consecutive keyframes $\text{b}_{k}$ and $\text{b}_{k+1}$, we have the following relationships:
\begin{equation}\small\label{eq:preintegration_p}
\Delta \textbf{p}_{k, k+1} = \textbf{R}^{\text{b}_{k}}_{\text{w}}(\text{s}(\bar{\textbf{p}}^{\text{w}}_{\text{b}_{k+1}} - \bar{\textbf{p}}^{\text{w}}_{\text{b}_k}) - \textbf{v}^{\text{w}}_{\text{b}_k}\Delta\textbf{t}_{k,k+1} - \frac{1}{2}\textbf{g}^{\text{w}}\Delta\textbf{t}^2_{k,k+1}),
\end{equation}
\begin{equation}\small\label{eq:preintegration_v}
\Delta \textbf{v}_{k, k+1} = \textbf{R}^{\text{b}_k}_{\text{w}}(\textbf{v}^{\text{w}}_{\text{b}_{k+1}} - \textbf{v}^{\text{w}}_{\text{b}_k}- \textbf{g}^{\text{w}}\Delta\textbf{t}_{k,k+1}),
\end{equation}
\begin{equation}\small\label{eq:camera2body}
\text{s}\bar{\textbf{p}}^{\text{w}}_{\text{b}_k} = \text{s}\bar{\textbf{p}}^{\text{w}}_{\text{c}_k} - \textbf{R}^{\text{w}}_{\text{b}_k}\textbf{p}^{\text{b}}_{\text{c}},
\end{equation}
where $\Delta \textbf{p}_{k, k+1}$ and $\Delta \textbf{v}_{k, k+1}$ are the preintagration of translation and velocity respectively from $k$-th to $k+1$-th keyframes.
We combine~\eq{eq:preintegration_p}$\sim$~\eq{eq:camera2body} into the following linear equation:
\begin{align}\small\label{eq:generallineareq}
\mathcal{A}_{k, k+1} \mathcal{X}_1 = \mathcal{B}_{k, k+1},
\end{align}
\begin{equation*}\small
\begin{split}
&\mathcal{A}_{k, k+1} = \\
&\left [ 
\begin{matrix}
0_{3\times 3k}&\boldsymbol{\alpha}^a_{k,k+1} & 0_{3\times 3} &0_{3\times 3(N-k-2)}&\boldsymbol{\alpha}^b_{k,k+1} & \boldsymbol{\alpha}^c_{k,k+1}\\
0_{3\times 3k}&\boldsymbol{\beta}^a_{k,k+1} & \boldsymbol{\beta}^b_{k,k+1}&0_{3\times 3(N-k-2)}&\boldsymbol{\beta}^c_{k,k+1} & 0_{3\times 1}
\end{matrix}
 \right],   
\end{split}
\end{equation*}
\begin{align*}\small
\mathcal{B}_{k, k+1} =
 \left [ 
\begin{matrix}
\Delta \textbf{p}_{k, k+1} - \textbf{p}^{\text{b}}_{\text{c}} + {\textbf{R}^{\text{b}_k}_{\text{w}}}\textbf{R}^{\text{w}}_{\text{b}_{k+1}}\textbf{p}^{\text{b}}_{\text{c}}\\
\Delta \textbf{v}_{k, k+1}\\
\end{matrix}
 \right],
\end{align*}
\begin{align*}\small
\begin{array}{ll}
\boldsymbol{\alpha}^a_{k,k+1} = -\textbf{R}^{\text{b}_k}_{\text{w}}\Delta\textbf{t}_{k,k+1},& 
\boldsymbol{\alpha}^b_{k,k+1} = -\frac{1}{2}\textbf{R}^{\text{b}_k}_{\text{w}}\Delta\textbf{t}^2_{k,k+1}, \\
\boldsymbol{\alpha}^c_{k,k+1} = \textbf{R}^{\text{b}_k}_{\text{w}}(\bar{\textbf{p}}^{\text{w}}_{\text{c}_{k+1}} - \bar{\textbf{p}}^{\text{w}}_{\text{c}_k}), &
\boldsymbol{\beta}^a_{k,k+1} = -\textbf{R}^{\text{b}_k}_{\text{w}},\\
\boldsymbol{\beta}^b_{k,k+1} = \textbf{R}^{\text{b}_k}_{\text{w}}, &
\boldsymbol{\beta}^c_{k,k+1} = -\textbf{R}^{\text{b}_k}_{\text{w}}\Delta\textbf{t}_{k,k+1},
\end{array}
\end{align*}
where the $\mathcal{A}_{k, k+1}$ matrix has dimensions ${6\times (3N+4)}$ and $\mathcal{B}_{k, k+1}$ is a ${6\times 1}$ vector. The camera's up-to-scale translations at two consecutive keyframes, $\bar{\textbf{p}}^{\text{w}}_{\text{c}_k}$ and $\bar{\textbf{p}}^{\text{w}}_{\text{c}_{k+1}}$, are obtained from the pure monocular SLAM, as well as the orientations of the IMU with respect to the world frame, $\textbf{R}_{\text{b}_{k}}^{\text{w}}$ and $\textbf{R}_{\text{b}_{k+1}}^{\text{w}}$. It is assumed that the extrinsic 
calibration
matrix $[\textbf{R}^{\text{b}}_{\text{c}} | \textbf{p}^{\text{b}}_{\text{c}}]$ is known, which allows for the transformation of camera poses to the IMU frame of reference. $\textbf{R}^{\text{b}_{k}}_{\text{w}}$ is the transpose of $\textbf{R}_{\text{b}_{k}}^{\text{w}}$. 

We then obtain $\mathcal{X}_1$ by considering all relationships among $N$ keyframes and solving the linear least squares problem:
 \[\min_{\mathcal{X}_1}\sum_{k\in \mathcal{K}}\left\|\mathcal{A}_{k, k+1}\mathcal{X}_1 - \mathcal{B}_{k, k+1}\right\|^2\]
where $\mathcal{K}$ indexes all $N$ keyframes.

\subsection{Refinement}
This step aims to obtain refined estimates of the keyframes’ velocities, gravity, and the scale factor from the previous step and to estimate the acceleration bias. The parameters that we want to estimate in this step are:
\begin{equation*}\small
\mathcal{X}_2 = \left [\textbf{v}^{\text{w}}_{\text{b}_{0}:\text{b}_{N-1}}, \ \textbf{b}_\text{a}, \ w_1, \ w_2, \ \text{s} \right]^\top.
\end{equation*}

As previously noted in \cite{weibovio}, distinguishing between acceleration bias and gravity can be challenging as they tend to be coupled and difficult to separate. As a result, VINS-Mono disregards acceleration bias during initialization and assumes it to be zero. Other methods, such as \cite{viorb}, rely on waiting for a prolonged period of time to observe these values. In our approach, we aim to decouple them by refining the initial gravity estimate from the previous step in its tangent space, and adding a weight matrix $\mathcal{W}$ to ~\eq{eq:generallineareq2} to keep acceleration bias at zero and handle outliers when the motion performed does not provide enough information or is blurred.


We refine the gravity estimate using an approach similar to VINS-Mono, which maintains the magnitude of the gravity vector and adjusts it with two variables in its tangent space. We also decouple the acceleration bias during this process. This allows us to represent the gravity vector in a more accurate way:
\begin{equation}\small\label{refineg}
\textbf{g}^{\text{w}} = g\textbf{g}^{\text{w}}_{unit} + {\delta \textbf{g}}, \ \ \delta \textbf{g} = w_1\textbf{b}_1+w_2\textbf{b}_2,
\end{equation}
where $g$ is the known magnitude of gravity, $\textbf{g}^{\text{w}}_{unit}$ is a unit vector denoting the gravity direction obtained by the previous step. $\textbf{b}_1$ and $\textbf{b}_2$ are two orthogonal basis vectors spanning the tangent plane. The initial values of $w_1$ and $w_2$ are set to zero. 

By substituting the value of $\textbf{g}^{\text{w}}$ from \eq{refineg} into \eq{eq:preintegration_p} and \eq{eq:preintegration_v}, and introducing the acceleration bias term by approximating the first order of $\Delta \textbf{p}_{k, k+1}$ and $\Delta \textbf{v}_{k, k+1}$, we can rewrite \eq{eq:generallineareq} to obtain a new equation with a weight matrix $\mathcal{W}_{k, k+1}$ between two consecutive keyframes $\text{b}_{k}$ and $\text{b}_{k+1}$ as follows:
\begin{align}\small\label{eq:generallineareq2}
\mathcal{W}_{k, k+1}\mathcal{H}_{k, k+1} \mathcal{X}_2 = \mathcal{W}_{k, k+1}\mathcal{Z}_{k, k+1}.
\end{align}
\begin{align*}\small
\mathcal{W}_{k, k+1} = \begin{bmatrix}
   w^{\alpha}_{k,k+1} & {0}_{3\times3} \\ 
   {0}_{3\times3} & w^{\beta}_{k,k+1}
 \end{bmatrix},
\end{align*}
\begin{align*}\small
\begin{array}{ll}
\textbf{e}_{\boldsymbol\alpha} = \mathcal{H}_{k,k+1}[0:2]\mathcal{X}_2-\mathcal{Z}_{k,k+1}[0:2], \\
\textbf{e}_{\boldsymbol\beta} = \mathcal{H}_{k,k+1}[3:5]\mathcal{X}_2-\mathcal{Z}_{k,k+1}[3:5],
\end{array}
\end{align*}
\begin{align*}\small
\begin{array}{ll}
w^{\alpha}_{k,k+1} = \text{diag}(\exp(-\Vert\textbf{e}_{\boldsymbol\alpha}\Vert),\exp(-\Vert\textbf{e}_{\boldsymbol\alpha} \Vert),\exp(-\Vert\textbf{e}_{\boldsymbol\alpha}\Vert)),\\
w^{\beta}_{k,k+1} = \text{diag}(\exp(-\Vert\textbf{e}_{\boldsymbol\beta}\Vert),\exp(-\Vert\textbf{e}_{\boldsymbol\beta}\Vert),\exp(-\Vert\textbf{e}_{\boldsymbol\beta}\Vert)),
\end{array}
\end{align*}
where $\mathcal{W}_{k, k+1}$ is a $6\times6$ matrix, $\textbf{e}_{\boldsymbol\alpha}$ and $\textbf{e}_{\boldsymbol\beta}$ are $3\times1$ vectors. $\mathcal{H}_{k,k+1}[i:j]$ denotes the submatrix of $\mathcal{H}_{k,k+1}$ from row $i$ to row $j$, and $\mathcal{Z}_{k,k+1}[i:j]$ denotes the submatrix of $\mathcal{Z}_{k,k+1}$ from row $i$ to row $j$. The detailed forms of matrices $\mathcal{H}_{k,k+1}$ and $\mathcal{Z}_{k,k+1}$ are given in the appendix.

By utilizing the solution of keyframes' velocities, scale, and gravity from the previous step as a seed, we can obtain $\mathcal{X}_2$ by solving the following linear least squares with preconditioned conjugate gradient (PCG):
 \[\min_{\mathcal{X}_2}\sum_{k\in \mathcal{K}}\left\|\mathcal{W}_{k, k+1}\mathcal{H}_{k, k+1}\mathcal{X}_2 - \mathcal{W}_{k, k+1}\mathcal{Z}_{k, k+1}\right\|^2\]
where $\mathcal{K}$ indexes all $N$ keyframes.

\section{Experimental Results} \label{sec:results}
We evaluate the proposed initialization method using the EuRoC dataset \cite{Burri25012016} and compare it to two state-of-the-art disjoint visual-inertial initialization methods: 1) a linear loosely-coupled method in VINS-Mono~\cite{VINS}, and 2) a non-linear optimization method in ORB-SLAM3~\cite{inertialonlyinit},\cite{ORBSLAM3TRO}. The evaluation metrics include computation speed, accuracy, and robustness. Note that since, according to Campos et al.~\cite{inertialonlyinit}, disjoint initialization generally outperforms joint approaches, we compare our method to disjoint  methods only.
We run each sequence five times, select the run that achieves median accuracy, and use that as the final outcome for all the metrics.
Overall, the proposed method achieves the best performance in terms of accuracy and robustness, at competitive computation speed.

\subsection{Experimental Setup}
The EuRoC dataset provides accurate rotation and translation data for 11 sequences recorded by a Micro Aerial Vehicle (MAV). These sequences vary from slow flights under favorable visual conditions to dynamic flights under challenging conditions such as motion blur, poor illumination, and occlusion. The dataset features visual-inertial sensor units that are hardware time synchronized, including: 1) two global shutter, monochrome cameras recording at 2x20 FPS and 2) a MEMS IMU providing angular rate and acceleration data at 200 Hz. Additionally, the dataset includes camera intrinsic and camera-IMU extrinsic parameters.  

To guarantee a fair comparison among the various initialization methods, the initialization part of VINS-Mono and EDI are integrated into ORB-SLAM3, enabling the evaluation of all methods using the same ORB-SLAM3 framework. All experiments are conducted on an Intel i7-10700K desktop with 32GB of RAM. 

Specifically, EDI and the initialization method of VINS-Mono are integrated into 
the Local Mapping thread in ORB-SLAM3, without affecting the real-time performance of the tracking thread.
The following parameters have been defined for EDI: the standard deviation of angular velocity measurement noise is set to 1.7$e^{-4}$ [rad/s], the standard deviation of angular velocity random walk noise is $2e^{-5} [rad/s\sqrt{s}]$, the magnitude of gravity is 9.81 [m/$s^2$] and the number of iterations in the PCG is 4.

\begin{table}[!b]
\centering
\caption{
Computation time for estimating inertial parameters for ORB-SLAM3~\cite{ORBSLAM3TRO}, VINS-Mono~\cite{VINS}, and our method (EDI) during initialization. Our approach, EDI, is an inertial only initialization method that estimates scale, keyframe velocities, gravity direction, and IMU biases using only inertial residuals, without considering visual residuals.}
 \resizebox{\columnwidth}{!}{
\begin{tabular}{@{}lc|cc|c}
\toprule  
\multirow{3}{*}{\textbf{Seq name}} &  \multicolumn{1}{c}{\textbf{EDI}} & \multicolumn{2}{c}{\textbf{ORB-SLAM3}}  & \multicolumn{1}{c}{\textbf{VINS-Mono}} \\
& \multicolumn{1}{c}{Time (ms)}  &\multicolumn{2}{c}{Time (ms)} &\multicolumn{1}{c}{Time (ms)}  \\
&Inert. Only &Inert. Only & Inert. Only + VI-BA & Inert. Only \\
  \hline
    V1\_01\_easy &0.44  &1.23 &60.30  & \textbf{0.30} \\
    V1\_02\_medium &0.46  & 1.55 &28.33&\textbf{0.22}   \\
    V1\_03\_difficult &\textbf{0.46}  &1.43 &41.24 &0.48 \\
  \hline
     V2\_01\_easy &0.53 & 1.20 &49.79 &\textbf{0.52}\\
    V2\_02\_medium &0.66 & 1.43 &41.22&\textbf{0.33} \\
    V2\_03\_difficult &\textbf{0.54} &1.16 &46.05 &0.66 \\
  \hline
    MH\_01\_easy & 0.60 &1.56 &22.50 &\textbf{0.52}\\
    MH\_02\_easy & \textbf{0.38}&1.22 &28.27 & 0.64\\
    MH\_03\_medium & \textbf{0.48} &1.13 &32.37 &\textbf{0.48} \\
    MH\_04\_difficult & 0.62 &1.50 &26.71&\textbf{0.44} \\
    MH\_05\_difficult  &0.75  &1.51 &22.25 & \textbf{0.31}\\
  \hline
    Avg &0.54 &1.36 &36.28 &  \textbf{0.45}\\
\bottomrule
\end{tabular}
}
\label{tab:init_time_result}
\end{table}

\subsection{Computation Speed Evaluation} \label{sec: speed performance}
A fast initialization is important to allow the SLAM system to proceed to the tracking step in real-time. EDI eliminates the need for the computationally intensive step of visual-inertial bundle adjustment (VI-BA) during the initialization, similar to VINS-Mono~\cite{VINS}, making it more efficient and faster.
Table~\ref{tab:init_time_result} presents the runtime comparison for estimating inertial parameters during the initialization stage involving 10 keyframes, with the best results for each sequence highlighted in bold.
The results show that our method is 2-3 times faster than the ORB-SLAM3~\cite{ORBSLAM3TRO} inertial-only method on average and has a comparable computational efficiency to VINS-Mono.
\subsection{Accuracy Evaluation} \label{sec: accu performance}
To measure accuracy, we evaluate both scale and trajectory error. 
Scale error measures how closely the estimated scale factor aligns with the true scale, calculated as $|s^* - \hat{s}|/|s^*| \times 100 \%$ , where $\hat{s}$ is the scale factor determined by aligning the estimated trajectory to the ground truth, and $s^*$ is 1. 

Table~\ref{tab:init_scale_ate_result} compares the scale error using 10 keyframe trajectories for initialization. Our method consistently achieves the highest accuracy, with a scale error of 5\% or less for most tasks. In contrast, the other two methods struggle to achieve similar levels of accuracy. Although ORB-SLAM3 with VI-BA performs better on the MH\_04\_difficult task, it takes more than 67 times longer to run on average compared to our method. Furthermore, the effectiveness of using VI-BA in ORB-SLAM3 heavily depends on the quality of the inertial-only estimation used as a seed for finding the optimal solution, and the initialization method in ORB-SLAM3 is highly sensitive to the scale factor.
As shown in Table~\ref{tab:traj_results}, our method's trajectory accuracy is comparable to ORB-SLAM3, with an average scale error of less than 1\% for entire trajectories after undergoing two rounds of VI-BA refinement, similar to what is used in ORB-SLAM3.

\begin{table}[!tbp]
\centering
{
\vspace{0.20in}
\caption{
Scale error comparison after initialization without VI-BA. The results of ORB-SLAM3 and VINS-Mono were obtained by executing the publicly available code with its default configuration. (Campos et al. \cite{inertialonlyinit} report the runs with the highest accuracy, while we report the median of five runs.)
}
\label{tab:init_scale_ate_result}
 \resizebox{\columnwidth}{!}{
\begin{tabular}{@{}lc|cc|c}
\toprule  
\multirow{3}{*}{\textbf{Seq name}} &  \multicolumn{1}{c}{\textbf{EDI}} & \multicolumn{2}{c}{\textbf{ORB-SLAM3}}  & \multicolumn{1}{c}{\textbf{VINS-Mono}} \\
&\multicolumn{1}{c}{Scale Error(\%)} & \multicolumn{2}{c}{Scale Error(\%)} &\multicolumn{1}{c}{Scale Error(\%)} \\
&Inert. Only & Inert. Only & Inert. Only + VI-BA & Inert. Only \\

  \hline
    V1\_01\_easy &\textbf{0.3}&2.9 &1.4&19.5  \\
    V1\_02\_medium &\textbf{0.4} &9.6&4.6 &18.9 \\
    V1\_03\_difficult &\textbf{9.8}&53.6&13.3& 14.9\\
  \hline
     V2\_01\_easy &\textbf{0.6} &2.8 &2.2 & 5.6\\
    V2\_02\_medium &4.4 &3.8 &3.8 & \textbf{3.4}\\
    V2\_03\_difficult &4.8 &7.7& \textbf{4.3}&21.9\\
  \hline
    MH\_01\_easy &3.3 &5.8&\textbf{3.0} &7.2  \\
    MH\_02\_easy &5.6&\textbf{5.4}&10.4 & 5.8 \\
    MH\_03\_medium &\textbf{3.4}&129.4 &158.5&23.7\\
    MH\_04\_difficult &19.9 &17.6 & \textbf{1.7}&13.3 \\
    MH\_05\_difficult  &\textbf{14.2}  &1017.4  & 97.6& 31.2\\
  \hline
    Avg &\textbf{5.8} &114.2 &27.3 &15.0\\
\bottomrule
\end{tabular}}
}
\end{table}

\begin{table}[!t]

\centering
{
\vspace{0.20in}
\caption{
Scale and full trajectory error comparison using different initialization methods after VI-BA. ATE(m) is the absolute trajectory error of the entire trajectory 
in meters without a Sim(3) transformation.
 }
\label{tab:traj_results}
\resizebox{\columnwidth}{!}{
\begin{tabular}{@{}lcccccc}
\toprule  
\multirow{2}{*}{\textbf{Seq name}} &  \multicolumn{2}{c}{\textbf{EDI}} &
\multicolumn{2}{c}{\textbf{ORB-SLAM3}} &
\multicolumn{2}{c}{\textbf{VINS-Mono}} \\
 & Scale Error(\%) & ATE(m) & Scale Error(\%) & ATE(m) & Scale Error(\%) & ATE(m)\\
 \toprule
 V1\_01\_easy &\textbf{0.9} & \textbf{0.031} &1.1& 0.032 &1.9  &0.044 \\
 V1\_02\_medium & \textbf{0.0} & \textbf{0.061}  &0.6 & 0.064 &0.6   &0.075  \\
 V1\_03\_difficult & 3.3   & \textbf{0.076} & 3.7   & \textbf{0.076} &\textbf{1.0} &0.103  \\
  \hline
 V2\_01\_easy & \textbf{1.1} & \textbf{0.059} & 1.5 &  0.060 &1.8  &0.063 \\
 V2\_02\_medium  & \textbf{0.1} & 0.063 &  0.8 &  \textbf{0.059} &0.7  &0.065 \\
 V2\_03\_difficult & \textbf{0.2} & \textbf{0.060} & 0.7&  0.063 &0.3& 0.075\\
  \hline
MH\_01\_easy &\textbf{1.5} & \textbf{0.085} & 2.2 & 0.093& 2.2   & 0.104 \\
MH\_02\_easy  & \textbf{0.3}& \textbf{0.077}  & 1.6 &0.081 &  1.0&0.081   \\
MH\_03\_medium & \textbf{0.4}& \textbf{0.066}& 0.5 & 0.070 &1.3  &0.198  \\
MH\_04\_difficult & \textbf{0.0} & 0.132 & 0.7 & \textbf{0.107}& 1.3  &  0.183\\
MH\_05\_difficult & 1.0 & 0.126 &1.3 &\textbf{0.110}  & \textbf{0.7}& 0.382\\
  \hline
Avg & \textbf{0.8} & 0.076  &1.3 & \textbf{0.074}& 1.3 & 0.125\\
\bottomrule
\end{tabular}
}
}
\end{table}

\subsection{Robustness Evaluation} \label{sec: robust performance}
A robust initialization step is crucial for ensuring a reliable starting point for the SLAM system to build upon. To evaluate the robustness of our method, we test it in challenging conditions such as motion blur and illumination change using sequences MH\_04\_difficult, MH\_05\_difficult, V1\_03\_difficult, and V2\_03\_difficult. Additionally, to make the task more challenging, we introduce noise to the rotation estimates of keyframes used in the initialization, obtained from a monocular SLAM system. 
The added noise has a standard deviation of 0.1 radians for roll, pitch, and yaw, and simulates a scenario in which pose estimation from pure monocular SLAM is of poor quality and the initialization methods are challenged to maintain accuracy and stability.

\begin{figure}
\vspace{0.08in}
\footnotesize
    \centering
    \begin{tabular}{cc}
        \includegraphics[width=0.4\columnwidth]{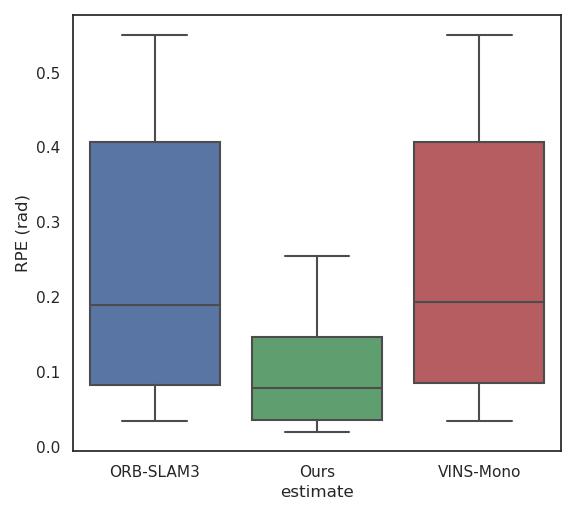}&
        \includegraphics[width=0.4\columnwidth]{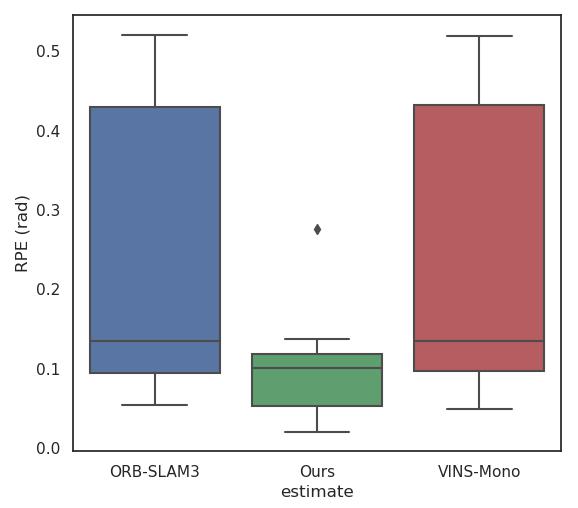}\\
        \hspace{0.2in} MH\_04\_difficult & \hspace{0.2in} MH\_05\_difficult \\
        \includegraphics[width=0.4\columnwidth]{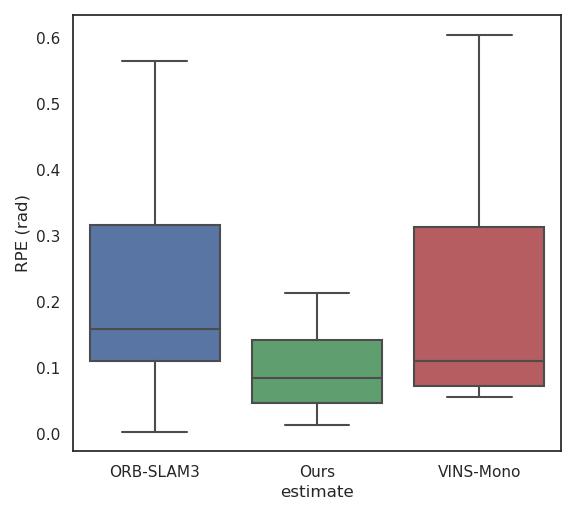}&
        \includegraphics[width=0.4\columnwidth]{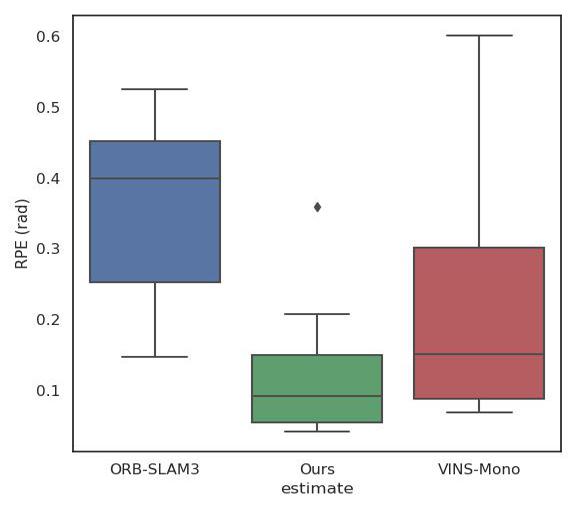}\\
        \hspace{0.2in} V1\_03\_difficult  & \hspace{0.2in} V2\_03\_difficult 
    \end{tabular}
    \caption{
    Box plots of the relative rotation error: comparison between our initialization method and ORB-SLAM3 for pure monocular SLAM’s rotation estimation with added noise.}
    \label{fig:rot_noise}
\end{figure}

As shown in Table~\ref{tab:rotation_noise} and \fig{fig:rot_noise}, our method outperforms ORB-SLAM3~\cite{ORBSLAM3TRO} and VINS-Mono~\cite{VINS} in rotation estimation, with a median error of 0.099-0.128 radians compared to ORB-SLAM3's 0.160-0.350 radians and VINS-Mono's 0.231-0.247 radians. It also has a lower root mean square error (rmse) of 0.116-0.127 radians compared to ORB-SLAM3's 0.297-0.374 radians and VINS-Mono's 0.291-0.304 radians in four challenging sequences with added noise. 
Our method, which uses ESKF during the initialization phase, is able to improve the accuracy of the rotation estimates, particularly when the visual estimates from a pure monocular SLAM system are not accurate.
In terms of the full trajectory, in one of the four tasks (V1\_03\_difficult sequence), our method can run the entire trajectory with an Absolute Trajectory Error (ATE) of 0.253 meters, while the other two methods fail to run all of the four challenging sequences with added noise, which demonstrates the robustness of our proposed method in comparison to the others.
\begin{table}[!tbp]
\vspace{0.05in}
\centering
\caption{
Relative rotation error comparison after the initialization in four challenging sequences with added noise.
The rotation error is calculated as the root mean square error of the relative rotation (RMSE).}
\scalebox{0.8}{
\begin{tabular}{@{}lccc}
\toprule  
\multirow{2}{*}{\textbf{Seq name}} &  \multicolumn{1}{c}{\textbf{EDI}} & \multicolumn{1}{c}{\textbf{ORB-SLAM3}}  & \multicolumn{1}{c}{\textbf{VINS-Mono}} \\
& RMSE (rad) &RMSE (rad) & RMSE (rad) \\
  \hline
    V1\_03\_difficult &\textbf{0.116} & 0.302 & 0.298 \\
    V2\_03\_difficult &\textbf{0.160} & 0.374 & 0.291 \\
    MH\_04\_difficult &\textbf{0.125} & 0.304 & 0.304 \\
    MH\_05\_difficult &\textbf{0.127} & 0.297 & 0.297\\
  \hline
    Avg &\textbf{0.132} &0.319 &0.298\\
\bottomrule
\end{tabular}}
\label{tab:rotation_noise}
\end{table}

\section{CONCLUSIONS}
Our proposed approach, EDI addresses the limitations of previous disjoint methods by utilizing an Error-state Kalman Filter (ESKF) to estimate gyroscope bias and correct rotation estimates, providing adaptability for application to other SLAM systems that use sensors such as GNSS and GPS. In addition, EDI offers a closed-form solution and introduces weights to handle outliers when estimating initial velocity, scale, gravity, and acceleration bias. EDI outperforms previous disjoint methods in terms of accuracy and robustness, at competitive computation speed, even in challenging environments with artificial noise. This new approach has promising potential for the development of efficient and reliable navigation systems in the future.
\section{APPENDIX}
Details of~\eq{eq:generallineareq2}
\begin{equation*}\small
\begin{split}
&\mathcal{H}_{k, k+1} = \\
&\left [ 
\begin{matrix}
\boldsymbol{\alpha}^a&\boldsymbol{\alpha}^b & \boldsymbol{\alpha}^c &\boldsymbol{\alpha}^d&\boldsymbol{\alpha}^e & \boldsymbol{\alpha}^f & \boldsymbol{\alpha}^g \\
\boldsymbol{\beta}^a&\boldsymbol{\beta}^b & \boldsymbol{\beta}^c&\boldsymbol{\beta}^d&\boldsymbol{\beta}^e & \boldsymbol{\beta}^f& \boldsymbol{\beta}^g
\end{matrix}
 \right],    
\end{split}
\end{equation*}
\begin{align*}\small
\mathcal{Z}_{k,k+1} =
 \left [ 
 \begin{matrix}
\Delta \textbf{p}_{k,k+1} - \textbf{p}^{\text{b}}_{\text{c}} + \textbf{R}^{\text{b}_k}_{\text{w}}\textbf{R}^{\text{w}}_{\text{b}_{k+1}}\textbf{p}^{\text{b}}_{\text{c}} + \frac{1}{2}\textbf{R}^{\text{b}_k}_{\text{w}}\Delta\textbf{t}^2_{k,k+1}\textbf{g}_\textbf{0}\\
\Delta \textbf{v}_{k,k+1} + \textbf{R}^{\text{b}_k}_{\text{w}}\Delta\textbf{t}_{k,k+1}\textbf{g}_\textbf{0}
\end{matrix}
 \right],
\end{align*}
\begin{align*}\small
\begin{array}{ll}
\boldsymbol{\alpha}^a= 0_{3\times3k}, &\boldsymbol{\beta}^a = 0_{3\times3k}\\
\boldsymbol{\alpha}^b = -\textbf{R}^{\text{b}_k}_{\text{w}}\Delta\textbf{t}_{k,k+1} &
\boldsymbol{\beta}^b = -\textbf{R}^{\text{b}_k}_{\text{w}}
\\
\boldsymbol{\alpha}^c= 0_{3\times3} &\boldsymbol{\beta}^c = \textbf{R}^{\text{b}_k}_{\text{w}}\\
\boldsymbol{\alpha}^d = 0_{3\times3(N-k-2)}&
\boldsymbol{\beta}^d = 0_{3\times3(N-k-2)}\\
\boldsymbol{\alpha}^e = -\boldsymbol{J}^{\Delta\textbf{p}}_{\textbf{b}_\text{a}}&
\boldsymbol{\beta}^e = -\boldsymbol{J}^{\Delta\textbf{v}}_{\textbf{b}_\text{a}}\\
\boldsymbol{\alpha}^f = -\frac{1}{2}\textbf{R}^{\text{b}_k}_{\text{w}}\textbf{b}\Delta\textbf{t}^2_{k,k+1}&
\boldsymbol{\beta}^f = -\textbf{R}^{\text{b}_k}_{\text{w}}\textbf{b}\Delta\textbf{t}_{k,k+1}\\
\boldsymbol{\alpha}^g = \textbf{R}^{\text{b}_k}_{\text{w}}(\bar{\textbf{p}}^{\text{w}}_{\text{c}_{k+1}} - \bar{\textbf{p}}^{\text{w}}_{\text{c}_k})&
\boldsymbol{\beta}^g = 0_{3\times 1}
\end{array}
\end{align*}
where $\mathcal{H}_{k, k+1}$ has dimensions $6\times (3N+6)$ and $\mathcal{Z}_{k,k+1}$ is a $6\times1$ vector.
$\textbf{g}_{\textbf{0}} = g\textbf{g}^{\text{w}}_{unit}$, where $\textbf{g}^{\text{w}}_{unit}$ is the unit vector of the gravity in the world frame. The Jacobians $\boldsymbol{J}^{\Delta\textbf{p}}_{\textbf{b}_\text{a}}$ and $\boldsymbol{J}^{\Delta\textbf{v}}_{\textbf{b}_\text{a}}$ represent how the preintegration changes due to a small difference in bias estimation, and the vector of biases $\textbf{b} = [\textbf{b}_1, \textbf{b}_2]^{\top}$ includes two bias terms, $\textbf{b}_1$ and $\textbf{b}_2$. These bias terms are used to perturb the gravity vector, and are chosen to be two orthogonal basis vectors on the tangent plane.

\printbibliography

\end{document}